\documentclass[times,twocolumn,final]{elsarticle}
\usepackage{amsfonts}
\usepackage{}

\usepackage{prletters}
\usepackage{framed,multirow}

\usepackage{amssymb}
\usepackage{latexsym}
\usepackage{mathrsfs}
\usepackage{url}
\usepackage{xcolor}
\usepackage{booktabs}
\definecolor{newcolor}{rgb}{.8,.349,.1}

\journal{Pattern Recognition Letters}

\begin{document}

\ifpreprint
  \setcounter{page}{1}
\else
  \setcounter{page}{1}
\fi

\begin{frontmatter}

\title{Shape recognition by bag of skeleton-associated contour parts}

\author[ad1]{Wei \snm{Shen}}
\author[ad1]{Yuan \snm{Jiang}}
\author[ad1]{Wenjing \snm{Gao}}
\author[ad1]{Dan \snm{Zeng}\corref{cor1}}
\cortext[cor1]{Corresponding author:
  Tel.: +86-13122300551;}
\ead{dzeng_shu@outlook.com}
\author[ad2]{Xinggang \snm{Wang}}
\address[ad1]{Key Lab of Specialty Fiber Optics and Optical Access Networks, Shanghai University, Shanghai 200072, China}
\address[ad2]{School of Electronic Information and Communications, Huazhong University of Science and Technology, Wuhan 430074, China}

\begin{abstract}
Contour and skeleton are two complementary representations for shape recognition. However combining them in a principal way is nontrivial, as they are generally abstracted by different structures (closed string \emph{vs} graph), respectively. This paper aims at addressing the shape recognition problem by combining contour and skeleton according to the correspondence between them. The correspondence provides a straightforward way to associate skeletal information with a shape contour. More specifically, we propose a new shape descriptor, named \textbf{S}keleton-associated \textbf{S}hape \textbf{C}ontext (SSC), which captures the features of a contour fragment associated with skeletal information. Benefited from the association, the proposed shape descriptor provides the complementary geometric information from both contour and skeleton parts, including the spatial distribution and the thickness change along the shape part. To form a meaningful shape feature vector for an overall shape, the Bag of Features framework is applied to the SSC descriptors extracted from it. Finally, the shape feature vector is fed into a linear SVM classifier to recognize the shape. The encouraging experimental results demonstrate that the proposed way to combine contour and skeleton is effective for shape recognition, which achieves the state-of-the-art performances on several standard shape benchmarks.

\vspace{0.2cm}
\\

\textbf{Keywords}: Shape recognition, skeleton-associated contour parts, bag of features
\vspace{0.2cm}

\end{abstract}

%

\end{frontmatter}


\section{Introduction}
\label{intro}
\begin{figure}[!t]
\centering
\includegraphics[width=1.0\linewidth]{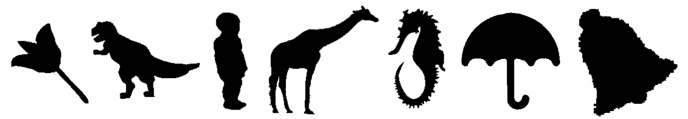}
\caption{Human biological vision system is able to recognize these object without any appearance information (brightness, color and texture).} \label{fig:shape_samples}
\end{figure}

Shape is a significant cue in human perception for object recognition. The objects shown in Fig.~\ref{fig:shape_samples} have lost their brightness, color and texture information and are only represented by their silhouettes, however it's not intractable for human to recognize their categories. This simple demonstration indicates that shape is stable to the variations in object color and texture and light conditions. Due to such advantages, recognizing objects by their shapes has been a long standing problem in the
literature. Shape recognition is usually considered as a classification problem that is given a testing shape, to determine its category label based on a set of training shapes as well as their category label. The main challenges in shape recognition are the large intra-class variations induced by deformation, articulation and occlusion. Therefore, the main focus of the research efforts have been made in the last decade~\cite{Ref:Belongie02,Ref:Lin07,Ref:Sun05,Ref:WangB10,Ref:Bai09,Ref:Bai14} is how to form a informative and discriminative shape representation.

Generally, the existing main stream shape representations can be classified into two classes: contour based~\cite{Ref:Belongie02,Ref:Lin07,Ref:Felzenszwalb07} and skeleton based~\cite{Ref:Aslan08,Ref:Bai08,Ref:Siddiqi99,Ref:Sebastian04,Ref:Xie08,Ref:ShenPR13}. The former one delivers the information that how the spatial distribution of the boundary points varies along the object contour. Therefore, it captures more informative shape information and is stable to affine transformation. However, it is sensitive to non-ridge deformation and articulation; On the contrary, the latter one provides the information that how thickness of the object changes along the skeleton. Therefore, it is invariant to non-ridge deformation and articulation, although it only carries more rough geometric features of the object. Consequently, such two representations are complementary. Nevertheless, very few works have tried to combine these two representations for shape recognition. The reason might be that combining the data of different structures is not trivial, as the contour is always abstracted by a closed string while the skeleton is abstracted either by a graph or a tree. Consequently, the matching methods~\cite{Ref:Duchon77,Ref:Demirci06,Ref:Cormen01,Ref:Ma15,Ref:Ma15d,Ref:Ma16,Ref:Ma14} for these two data abstraction are different. ICS~\cite{Ref:Bai09} is the first work to explicitly discuss how to combine contour and skeleton to improve the performance of shape recognition. However, the combination proposed in this work is just a weighted sum of the outputs of two generative models trained individually on contour features and skeleton features respectively. Therefore, how to combine contour and skeleton into a shape representation in a principled way is still an open problem.

\begin{figure}[!t]
\centering
\includegraphics[width=1.0\linewidth]{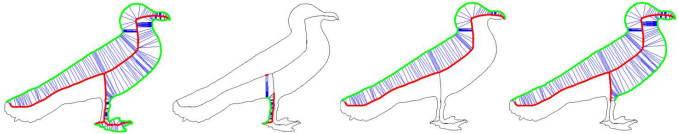}
\caption{Some corresponding contour and skeleton parts, marked in green and red, respectively. The corresponding contour and skeleton points are linked by blue lines.} \label{fig:contour-skeleton-parts}
\end{figure}

In this paper, our goal is to address the above combination issue to explore the complementarity between contour and skeleton to improve the performance of shape recognition. The main obstacle of the combination is the data structures of contour and skeleton are different (closed string \emph{vs} graph). A contour is usually described by the features of its parts (contour fragments)~\cite{Ref:Sun05,Ref:Felzenszwalb07}. As the correspondence between contour points and skeleton points can be obtained easily, for each contour point, we can associate the geometric information of its corresponding skeleton point with it. In this way, we can record the change of the object thickness, i.e., the skeleton radius, along each contour fragment. Such association actually leads to the combination of contour and skeleton on part level (Fig.~\ref{fig:contour-skeleton-parts} shows some corresponding contour and skeleton parts. Note that, a contour fragment may correspond to more than one skeleton segments, such as the second example in Fig.~\ref{fig:contour-skeleton-parts}). Therefore, combing contour and skeleton on part level is a feasible way.

\begin{figure}[!t]
\centering
\includegraphics[width=0.8\linewidth]{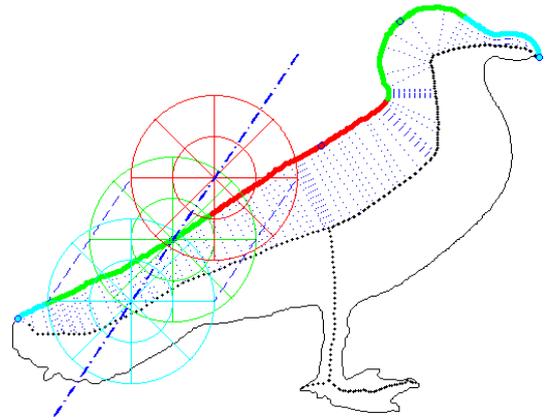}
\caption{The Skeleton-associated Shape Context descriptor of a contour point in a contour fragment, which is a 3D tensor to describe the Euclidean distances, orientations and thickness differences between the contour point and others in the fragment. It equals to the concatenated shape context descriptors~\cite{Ref:Belongie02} computed on sub-parts (marked by different colors) separated according to the object thickness differences between the contour point and others in the fragment.} \label{fig:SSC}
\end{figure}

With the extra information provided by skeleton, inspired by the well known descriptor \textbf{S}hape \textbf{C}ontext (SC) ~\cite{Ref:Belongie02}, we propose to encode the features of a contour point into a 3D tensor, in which the three dimensions describe the Euclidean distances, orientations and thickness differences between the contour points and others in the fragment, respectively. Intuitively, the proposed new descriptor extends SC by including the extra information, object thickness, provided by skeleton. Therefore, it is more informative; Essentially, this new descriptor is formed by concatenating the SC descriptors of the sub-parts of the contour fragment separated according to thickness information. Such sub-parts based representation capture fine level geometric information, so it is more discriminative. Fig.~\ref{fig:SSC} illustrates the new descriptor for a contour point in a contour fragment, in which the sub-parts of the contour fragment are marked by different colors and the sub-part and its SC descriptor are marked by the same color. This new shape descriptor is termed as \textbf{S}keleton-associated \textbf{S}hape \textbf{C}ontext (SSC), as it associates the skeletal information with the contour descriptor.

\begin{figure*}[!t]
\centering
\includegraphics[width=1.0\linewidth]{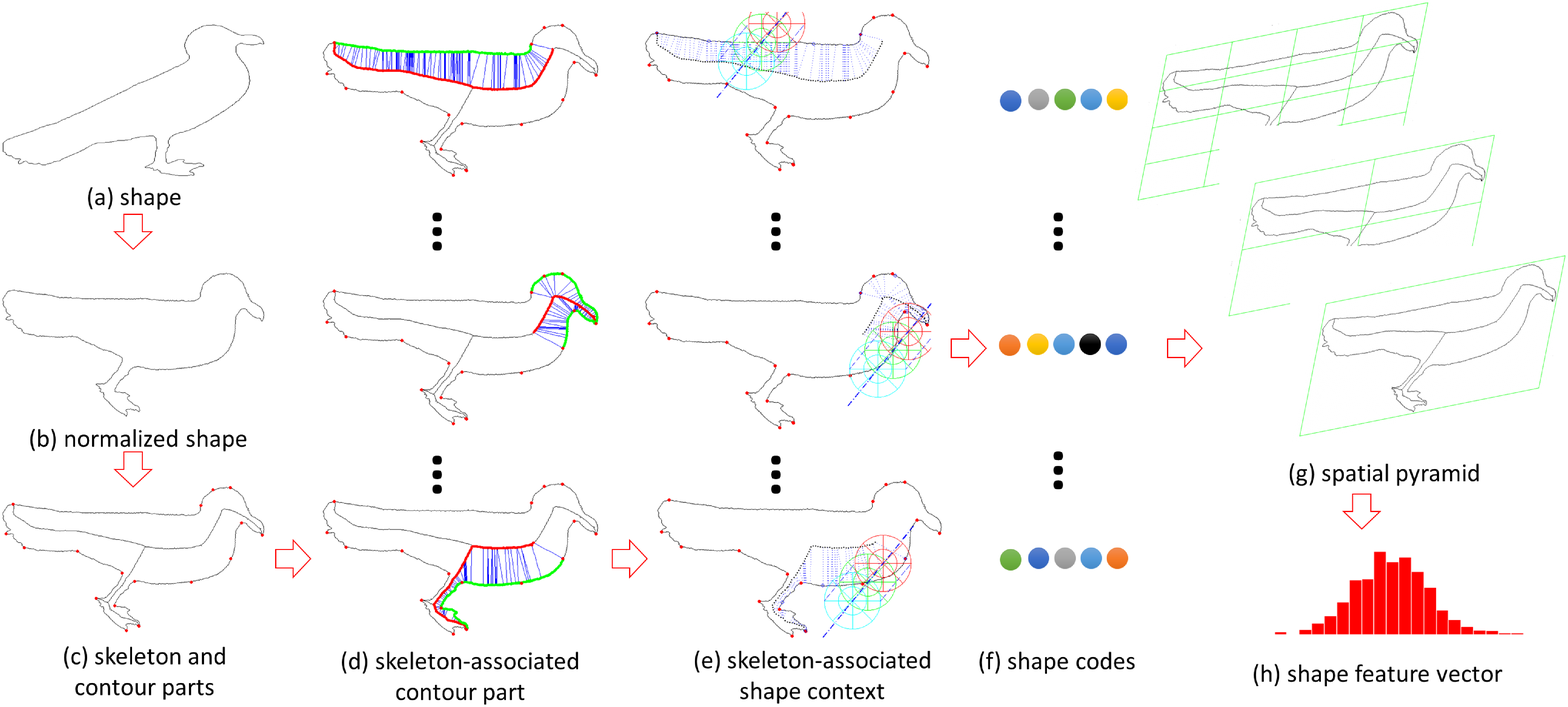}
\caption{The pipeline of building a shape feature vector by bag of skeleton-associated contour parts} \label{fig:pipeline}
\end{figure*}

Following the framework of the recent work \textbf{B}ag of \textbf{C}ontour \textbf{F}ragments (BCF)~\cite{Ref:Wang14}, we can obtain a shape feature vector of an overall shape by encoding and then pooling the SSC descriptors extracted from it. We term our method as \textbf{B}ag of \textbf{S}keleton-associated \textbf{C}ontour \textbf{P}arts (BSCP), as it associates skeletal information with contour fragments and encodes the shape features from shape part level. Fig.~\ref{fig:pipeline} shows the pipeline of building a shape feature vector by BSCP. Given a shape, firstly a normalization step is performed to align the shape according to its major axis (Fig.~\ref{fig:pipeline}(b)), as the \textbf{S}patial \textbf{P}yramid \textbf{M}atching (SPM)~\cite{Ref:Lazebnik06} step (Fig.~\ref{fig:pipeline}(g)) is not rotation invariant. Then, the skeleton of the shape is extracted and the contour of the shape is decomposed into contour fragments (Fig.~\ref{fig:pipeline}(c)). Each contour point is associated with a object thickness value, i.e, the radius of its corresponding skeleton point. A shape part is then described by the contour fragment associated with the object thickness values provided by its corresponding skeleton segments (Fig.~\ref{fig:pipeline}(d)). After that, each shape part is represented by concatenating the SSC descriptors extracted on its reference points (Fig.~\ref{fig:pipeline}(e)), and then encoded into shape codes (Fig.~\ref{fig:pipeline}(f)). To encode shape parts, we adopt local-constrained linear coding (LLC)~\cite{Ref:Wang10} scheme, as it has been proved to be efficient and effective for image classification. Finally, the shape codes are pooled into a compact shape feature vector by SPM (Fig.~\ref{fig:pipeline}(h)). The obtained shape feature vectors can be fed into any discriminative models, such as SVM and Random Forest, to perform shape classification. Using such discriminative models for shape recognition is more efficient than traditional shape classification methods, as the latter require time consuming matching and ranking steps.

Our contributions can be summarized in three aspects. First, we propose a natural way to associate a shape contour with skeletal information. Second, we propose a new shape descriptor which encodes the shape features from a contour fragment associated with skeletal information. Last, our method, Bag of Skeleton-associated Contour Parts achieves the state-of-the-arts on several shape benchmarks.

The remainder of this paper is organized as follows. Sec.~\ref{sec:related_work} reviews the works related to shape recognition. Sec.~\ref{sec:method} introduces the proposed shape descriptor as well as our framework for shape recognition. Experimental results and analysis on several shape benchmarks are shown in Sec.~\ref{sec:exp}. Finally, we draw the conclusion in Sec.~\ref{sec:con}.

Our preliminary work~\cite{Ref:Shen14} also combines contour and skeleton for shape recognition, while the difference to this paper is obvious. Rather than simply concatenating the contour and skeleton features on mid-level, this paper associates skeletal information with a shape contour on low-level by making full use of the natural correspondence between a contour and its skeleton.
\section{Related Work} \label{sec:related_work}
There have been a rich body of works concerning shape recognition in recent years~\cite{Ref:Belongie02,Ref:Lin07,Ref:Sun05,Ref:Grigorescu03,Ref:Baseski09,Ref:Daliri10,Ref:WangB10,Ref:Erdem10,Ref:Bicego15,Ref:Bharath15}. In the early age, the exemplar-based strategy has been widely used, such as~\cite{Ref:Belongie02,Ref:Lin07}. Generally, there are two key steps in this strategy. The first one is extracting informative and robust shape descriptors. For example, Belongie \emph{et al.}~\cite{Ref:Belongie02} introduce a shape descriptor named shape context (SC) which describes the relative spatial distribution (distance and orientation) of landmark points sampled on the object contour around feature points. Ling and Jacobs~\cite{Ref:Lin07} use inner distance to extend shape context to capture articulation. As for skeleton based shape descriptors, the reliability of them is ensured by effective skeletonization~\cite{Ref:Saha15,Ref:Borgefors99} or skeleton pruning~\cite{Ref:Bai07,Ref:Shen11} methods to a large extent. Among them, the shock graph and its variants~\cite{Ref:Siddiqi99,Ref:Sebastian04,Ref:Macrini11} are most popular, which are abstracted from skeletons by designed shape grammar. The second one is finding the correspondences between two sets of the shape descriptors by matching algorithms such as Hungarian, thin plate spline (TPS) and dynamic programming (DP). A testing shape is classified into the class of its nearest neighbor ranked by the matching costs. The exemplar-based strategy requires a large number of training data to capture the large intra-class variances of shapes. However, when the size of training set become quite large, it's intractable to search the nearest neighbor due to the high time cost caused by pairwise matching.

Generative models are also used for shape recognition. Sun and Super~\cite{Ref:Sun05} propose a Bayesian model, which use the normalized contour fragments as the input features for shape classification. Wang \emph{et al.}~\cite{Ref:WangB10} model shapes of one class by a skeletal prototype tree learned by skeleton graph matching. Then a Bayesian inference is used to compute the similarity between a testing skeleton and each skeletal prototype tree. Bai \emph{et al.}~\cite{Ref:Bai09} propose to integrate contour and skeleton by a Gaussian mixture model, in which contour fragments and skeleton paths are used as the input features. Unlike their method, ours encodes the contour and skeleton features into one shape descriptor according to the association between contour and skeleton. Therefore, we avoid the intractable step to finetune the weight between contour and skeleton models.

Recently, researchers begin to apply the powerful discriminative models to shape classification. Daliri and Torre~\cite{Ref:Daliri08,Ref:Daliri10} transform the contour into a string based representation according to a certain order of the corresponding contour points found during contour matching. Then they apply SVM to the kernel space built from the pairwise distances between strings to obtain classification results. Edem and Tari~\cite{Ref:Erdem10} transform a skeleton into a similarity vector, in which each element is the similarity between the skeleton and a skeletal prototype of one shape category. Then they apply linear SVM to the similarity vector to determine the category of the skeleton. Wang \emph{et al.}~\cite{Ref:Wang14} utilize LLC strategy to extract the mid-level representation BCF from contour fragments and they also use linear SVM for classification. Such coding based methods are used for 2D and 3D shape retrieval~\cite{Ref:Bai14,Ref:Bai15}. Shen \emph{et al.}~\cite{Ref:Shen14} propose a skeleton based mid-level representation named \textbf{B}ag of \textbf{S}keleton \textbf{P}aths (BSP), and concatenate the BCF and BSP for shape recognition. The weights between BCF and BSP are automatically learned by SVM. This method implicitly combines contour and skeleton according to the weights learned by SVM, while this paper explicitly combines contour and skeleton by using the correspondence between them, which is a more natural combination way.
\section{Methodology}\label{sec:method}
In this section, we will introduce our method for shape recognition, including the steps of shape normalization, SSC descriptor and shape classification by BSCP.
\subsection{Shape Normalization}
As the SPM strategy assumes that the parts of shapes falling in the same subregion are similar, it is not rotation invariant. To apply SPM to shape classification, a normalization step is required to align shapes roughly. One straightforward solution is to align each shape with its major axis. Here, we use principal component analysis (PCA) to compute the orientation of the major axis of each shape. Formally, given a shape $F\subset\mathbb{R}^2$, we apply PCA to the point set $\{p_i=(x_i,y_i)|p_i{\in}F\}_{i=1}^N$. First, the $N{\times}N$ covariance matrix $\Sigma$ is computed by $\Sigma = \frac{1}{N-1}\sum_{i=1}^N(x_i-\overline{x_i})(y_i-\overline{y_i})$, where $\overline{x_i}=\sum_{i=1}^Nx_i/N$ and $\overline{y_i}=\sum_{i=1}^Ny_i/N$. Then, the two eigenvectors $\mathbf{v}_1$ and $\mathbf{v}_2$ of $\Sigma$ form the columns of the $N{\times}N$ matrix $V$, and the two eigenvalues of $\Sigma$ are $(\lambda_1,\lambda_2)^{\mathrm{T}}=\textbf{diag}(V^{\mathrm{T}}{\Sigma}V)$. The orientation of the major axis of the shape $F$ is the orientation of the eigenvector whose corresponding eigenvalue is bigger. All shapes are rotated to ensure their estimated major axes are aligned with the horizontal line, such as the example given in Fig.~\ref{fig:pipeline}(b).
\subsection{Skeleton-associated Shape Context}
In this section, we show how to compute the SSC descriptor for a given contour point step by step.
\subsubsection{Skeleton-associated Contour}
For a given shape $F$, let $\mathcal {C}(F)$ and $\mathcal {S}(F)$ denote its contour and skeleton, respectively. The skeleton $\mathcal {S}(F)$ can be obtained by the method introduced in~\cite{Ref:Shen13}, which does not require parameter tuning for skeleton computation. Our goal is to find the corresponding skeleton point of each contour point and assign a object thickness value to it. To describe our method clearly, here we first briefly review some skeleton related definitions. According to the definition of skeleton~\cite{Ref:Blum73}, a skeleton is a set of the centers of the maximal discs of a shape. A maximal disc has at least two points of tangency on the contour, which are called \textbf{G}enerating \textbf{P}oints (GPs).

Formally, for a skeleton point $p (p{\in}\mathcal {S}(F))$, let $\mathscr{R}(p)$ be the radius of the maximal disc of the shape $F$ centered at $p$ and $\mathcal {G}(p)$ be the set of GPs of $p$. On the discrete domain, $\mathscr{R}(p)$ can be approached by the \textbf{D}istance \textbf{T}ransform (DT) value of $p$ to the contour $\mathcal {C}(F)$:
\begin{equation}
\mathscr{R}(p) = \min_{q{\in}\mathcal {C}(F)}\|p-q\|_2,
\end{equation}
where $\|\cdot\|_2$ is the $\ell_2$-Norm. $\mathcal {G}(p)$ can be obtained approximatively by
\begin{equation}
\mathcal {G}(p) = \{q^*|{\exists}{p_n{\in}\mathcal {N}(p)},q^*=\arg\min_{q{\in}\mathcal {C}(F)}\|q-p_n\|_2\},
\end{equation}
where $\mathcal {N}(p)$ denotes the eight neighbors of $p$. Note that, $\mathcal {G}(p){\subset}\mathcal {C}(F)$. Now we have a one-to-many correspondence between a skeleton point $p$ and a set of contour point $\mathcal {G}(p)$. For each contour point $q{\in}\mathcal {G}(p)$, we associate the object thickness value $\mathscr{R}(p)$ with it, and use the notation $  \mathscr{C}(\cdot)$ to denote the corresponding function mapping it to the skeleton point $p$, i.e., $p=\mathscr {C}(q)$, if $q{\in}\mathcal {G}(p)$. Now considering the overall shape, let $\mathcal {G}(S(F))$ be the set of all the GPs of $S(F)$:
\begin{equation}
\mathcal {G}(\mathcal {S}(F))=\bigcup_{p{\in}\mathcal {S}(F)}\mathcal {G}(p).
\end{equation}
Note that $\mathcal {G}(\mathcal {S}(F)){\subseteq}\mathcal {C}(F)$, so the function $ \mathscr{C}(\cdot)$ can not be applied to all the contour points. However, we can define a unified function $\mathscr{R}_a(\cdot)$ to compute the associated object thickness value for each contour point $q$:
\begin{equation}\label{eq:assoc}
\mathscr{R}_a(q)=\mathscr{R}(\mathscr{C}(q_a)),
\end{equation}
where $q_a = \arg\min_{q_g{\in}\mathcal {G}(\mathcal {S}(F)}l(q,q_g)$ and $l(\cdot,\cdot)$ is denoted by the minimum contour curve length between two contour points. Eq.~\ref{eq:assoc} means that for each contour point $q$, we search its closest contour point $q_a{\in}\mathcal {G}(\mathcal {S}(F))$ along the contour (if $q{\in}\mathcal {G}(\mathcal {S}(F)$, then $q_a=q$), and assign the associated object thickness value of $q_a$ to $q$.
\subsubsection{Shape Descriptor Computation}
Part-based methods~\cite{Ref:Sun05,Ref:Felzenszwalb07,Ref:Bai09,Ref:Wang14} have been widely used for shape recognition, as shape parts are the basic meaningful elements of a shape. We want to build a discriminative and informative shape representation based on shape parts. The shape parts can be obtained by any contour decomposition methods, such like \textbf{D}iscrete \textbf{C}ontour \textbf{E}volution (DCE)~\cite{Ref:LateckiL99}. Given a shape contour $\mathcal {C}(F)$, we apply DCE to obtain its critical points $\{u_i\}_{i=1}^T$, where $T$ is the number of the critical points. We build a shape part set $\mathcal {P}_{\mathcal {C}(F)}$, which consists of the contour fragments between any pairs of critical points $u_i,u_j$. Let $c_{ij}$ denote the contour fragment from $u_i$ to $u_j$ (anticlockwise direction), then we have
\begin{equation}
\mathcal {P}_{\mathcal {C}(F)} = \{c_{ij}|i{\neq}j,i,j{\in}\{1,\ldots,T\}\}.
\end{equation}
Note that we do not force $u_i$ and $u_j$ to be adjacent points in the critical point set, and $c_{ij}$ and $c_{ji}$ are two different parts. Also we have $\mathcal {C}(F)=c_{ij}{\bigcup}c_{ji}$. Using the method described in the previous section, any contour part can be transformed into a skeleton-associated contour part. In the reminder of this paper, unless otherwise specified, we treat these two concepts equally.

Now we propose how to compute the SSC descriptor at a reference contour point of a skeleton-associated contour part. Each point $q$ on a skeleton-associated contour part can be represented by a triplet $(x,y,\mathscr{R}_a(q))$, where $(x,y)$ is the relative coordinate and $\mathscr{R}_a(q)$ is the associated object thickness value{\footnote{To ensure scale invariant, this value should be normalized by dividing by the mean value of the points on the contour part.}}. From this view, the point $q$ actually lies in a 3D space. Given a contour part, we uniformly sample $n$ points on it, then for a given reference contour point $r_i$, we describe its descriptor by the distribution of relative differences to the $n$ sampled points on Euclidean distance, orientation and associated object thickness value. We compute a coarse histogram $h_i$ for $r_i$:
\begin{equation}
h_i(j)=\#\{q{\neq}r_i:(q-r_i){\in}{bin}(j)\},j{\in}\{1,\ldots,M\}
\end{equation}
Here, $(q-r_i)=(\rho_i,\theta_i,\log(\mathscr{R}_a(q)) - \log(\mathscr{R}_a(r_i)))$, where $\rho_i$ and $\theta_i$ are the Euclidean distance between $q$ and $r_i$ and the orientation angle of the ray from $r_i$ to $q$ defined on log-polar space, respectively. We use $M$ bins that are uniform in such a 3D space, which follows the strategy used in SC~\cite{Ref:Belongie02} to make the descriptor more sensitive to nearby sample points than those farther away. The histogram $h_i$ is defined to be the SSC of $r_i$.

Finally, we concatenate the SSC descriptors of the reference points on a contour part $c_{ij}$ to form the descriptor vector $\mathbf{f}_{ij}{\in}\mathbb{R}^D$ for $c_{ij}$: $\mathbf{f}_{ij}=(h_i;i=1,\ldots,n)^{\mathrm{T}}$, where $n$ is the number of the reference points and $D=n{\times}M$.
\subsection{Bag of Skeleton-associated Contour Parts}
In this section, we introduce how to perform shape classification by BSCP.
\subsubsection{Contour Parts Encoding}
Encoding a skeleton-associated contour part $\mathbf{f}{\in}\mathbb{R}^D$ is transforming it into a new space $\mathcal {B}$ by a given codebook with $K$ entries, $\mathbf{B} =(\mathbf{b_1},\mathbf{b_2},{\ldots},\mathbf{b_K})\in\mathbb{R}^{D{\times}K}$. In the new space, the contour part $\mathbf{f}$ is represented by a shape code $\mathbf{c}{\in}\mathbb{R}^K$.

Codebook construction is usually achieved by unsupervised learning, such as k-means. Given a set of contour parts randomly sampled from all the shapes in a dataset as well as their flipped mirrors, we apply k-means algorithm to cluster them into $K$ clusters and construct a codebook $\mathbf{B} =(\mathbf{b_1},\mathbf{b_2},{\ldots},\mathbf{b_K})$. Each cluster center forms an entry of the codebook $\mathbf{b_i}$.

To encode a contour part $\mathbf{f}$, we adopt LLC scheme~\cite{Ref:Wang10}, as it has been proved to be effective for image classification. Encoding is usually achieved by minimizing the reconstruction error. LLC additionally incorporates locality constraint, which solves the following
constrained least square fitting problem:
\begin{equation}
\min_{{\mathbf{c}}_{\pi_k}}\|\mathbf{f}-\mathbf{B}_{{\pi_k}}\mathbf{c}_{{\pi_k}}\|,~~\textbf{s.t.}~~\mathbf{1}^{\mathrm{T}}\mathbf{c}_{{\pi_k}}=1,
\end{equation}
where $\mathbf{B}_{\pi_k}$ is the local bases formed by the $k$ nearest neighbors of $\mathbf{f}$ and $\mathbf{c}_{\pi_k}{\in}\mathbb{R}^k$ is the reconstruction coefficients. Such a locality constrain leads to several favorable properties such as local smooth sparsity and better reconstruction. The code of $\mathbf{f}$ encoded by the codebook $\mathbf{B}$, i.e. $\mathbf{c}{\in}\mathbb{R}^K$, can be easily converted from $\mathbf{c}_{\pi_k}$ by setting the corresponding entries of $\mathbf{c}$ are equal to $\mathbf{c}_{\pi_k}$'s and others are zero.

\begin{figure}[!h]
\centering
\includegraphics[width=1.0\linewidth]{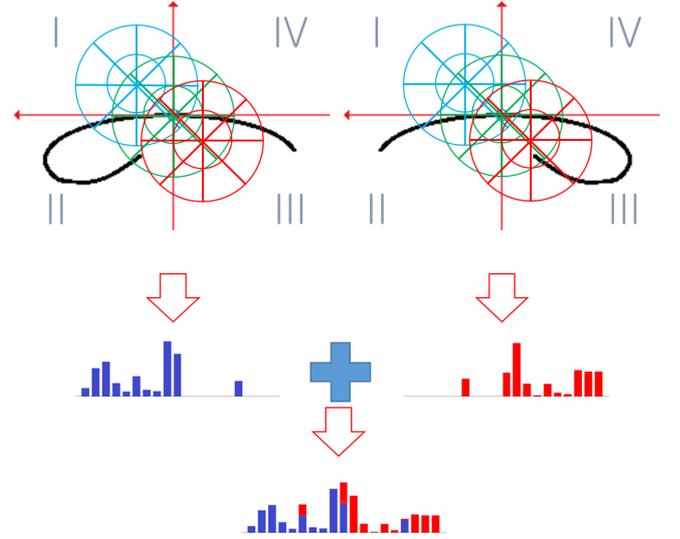}
\caption{The shape codes of a contour part and its flipped mirror are added in an element-wise manner to form the final shape code for it, which is invariant to flip transformation.} \label{fig:mirror}
\end{figure}

Note that, the SSC descriptors of a contour part and its flipped mirror are different, as shown in Fig.~\ref{fig:mirror}. To make our shape code invariant to the flip transformation, for a contour part, we propose to add the shape code of its flipped mirror to its in an element-wise manner (as shown in Fig.~\ref{fig:mirror}). In this way, the shape codes of a contour part and its flipped mirror are the same. The available encoding of contour parts and their flipped mirrors are ensured by the sufficient samples used for codebook building (recall that our codebook is generated by clustering a set of contour parts randomly sampled from all the shapes in a dataset as well as their flipped mirrors).
\subsubsection{Shape Code Pooling}
Given a shape $F$, its skeleton-associated contour parts are encoded into shape codes $\{\mathbf{c}_i\}_{i=1}^n$, where $n$ is the number of the contour parts in $F$. Now we describe how to obtain a compact shape feature vector by pooling the shape codes. SPM is usually used to incorporate spatial layout information when pooling the image codes. It usually divides a image into $2^l{\times}2^l (l=0,1,2)$ subregions and then the features in each subregion are pooled respectively. For the aligned shapes belong to one category, the contour parts falls in the same subregions should be similar. Here, the position of a contour part is defined as its median point. More specifically, we divide a shape $F$ into $2^l{\times}2^l (l=0,1,2)$ subregions, i.e. $21$ subregions totally. Let $\mathbf{c}^{z}{\in}\mathbb{R}^K$ denote the shape code of a contour part at position $z$, to obtain a shape feature vector $\mathbf{g}(F)$, for each subregion $SR_i,i{\in}(1,2,{\ldots},21)$, we perform max pooling on it as follow:
\begin{equation}
\mathbf{g}_i(F) = \max(\mathbf{c}^{z}|z{\in}SR_i),
\end{equation}
where the ``max'' function is performed in an element-wise manner, i.e. for each codeword, we take the max value of all shape codes in a subregion. Max pooling is robust to noise and has been successfully applied to image classification. $\mathbf{g}_i(F)$ is a $K$ dimensional feature vector of the subregion $SR_i$. The BSCP vector $\mathbf{g}(F)$ is a concatenation of the feature vectors of all subregions:
\begin{equation}
\mathbf{g}(F) = (\mathbf{g}^{\mathrm{T}}_1(F),\mathbf{g}^{\mathrm{T}}_2(F),{\ldots},\mathbf{g}^{\mathrm{T}}_{21}(F))^{\mathrm{T}}.
\end{equation}
Finally, $\mathbf{g}(F)$ is normalized by its $\ell_2$-norm: $\mathbf{g}(F)={\mathbf{g}(F)}/\|{\mathbf{g}(F)}\|_2$.
\subsection{Shape Classification by BSCP}
Given a training set $\{(\mathbf{g}_i, y_i)\}_{i=1}^M$ consisting of $M$ shapes from $L$ classes, where $\mathbf{g}_i$ and $y_i{\in}\{1,2,{\ldots},L\}$ are the BSCP vector and the class label of $i$-th shapes respectively, we train a multi-class linear SVM~\cite{Ref:Crammer01} as the classifier:
\begin{equation}
\min_{\mathbf{w}_1,{\ldots},\mathbf{w}_L}\sum_{j=1}^M\|\mathbf{w}_j\|^2+\alpha\sum_i\max(0,1+\mathbf{w}_{l_i}^{\mathrm{T}}\mathbf{g}_i-\mathbf{w}_{y_i}^{\mathrm{T}}\mathbf{g}_i),
\end{equation}
where $l_i=\arg\max_{l{\in}\{1,2,{\ldots},L\},l{\neq}y_i}\mathbf{w}_{l}^{\mathrm{T}}\mathbf{g}_i$ and $\alpha$ is a parameter to balance the weight between the regularization term (left part) and the multi-class hinge-loss term (right part). For a testing shape vector $\mathbf{g}$, its class label is given by
\begin{equation}
\widehat{y}=\arg\max_{l{\in}\{1,2,{\ldots},L\}}\mathbf{w}_{l}^{\mathrm{T}}\mathbf{g}.
\end{equation}

Here we adopt linear SVM, as the proposed BSCP feature vector is a high dimensional sparse vector, computed by LLC coding. The $\ell_2$ normalization in LLC makes the inner product of any vector with itself to be one, which is desirable for linear kernels~\cite{Ref:Wang10}. Using classifiers with nonlinear kernel, such as kernel SVM and random forest, instead leads to performance decrease.  
\section{Experimental Results}\label{sec:exp}
In this section, we evaluate our method on several shape benchmarks in comparison to the state-of-the-arts. We also investigate the effects of two important parameters introduced in our method on classification accuracy: the number of object thickness difference bins for computing SSC $N_{td}$ and codebook size $K$.

\begin{figure}[!h]
\centering
\includegraphics[width=1.0\linewidth]{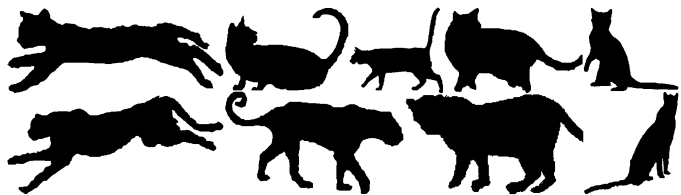}
\caption{Shapes of two classes from Animal dataset~\cite{Ref:Bai09}. The first row shows 5 shapes of the Cat class, with large intra-class variations caused by view point change and various gestures of the cats. Moreover, leopards on the second row are similar to those cats on the first row, which makes recognition of these two kinds of shapes much more difficult.} \label{fig:animal}
\end{figure}

\subsection{Experimental Setup}
For each contour part, we form a descriptor vector for it by concatenating the SSC descriptors computed on $5$ reference points. Unless otherwise specified, we set the number of bins for computing SSC to 300 (5 Euclidean distance bins, 12 orientation bins, 5 object thickness difference bins). Thus the dimension of a descriptor vector for a contour part is 1500. The number of Euclidean distance bins and the number of orientation bins are set to the default values used in SC~\cite{Ref:Belongie02}. Hence, we will discuss the effects of the number of object thickness difference bins on classification accuracy individually. When learning the codebook, the number of cluster centers (codebook size) is set to 2500 by default. We also study the performances of BSCP by varying the codebook size. To encode a contour part, we adopt the approximated LLC with $5$ nearest neighbors. When pooling, a shape is divided into $1\times1$, $2\times2$ and $4\times4$, in total 21 regions. The weight between the regularization term and the multi-class hinge-loss term in the multi-class linear SVM formulation is set to 10. Default parameter settings reported in~\cite{Ref:Shen13} are adopted to extract skeletons.

All the experiments were carried out on a workstation (3.1GHz 32-core CPU, 128G RAM and Ubuntu14.04 64-bit OS). It takes about $25$ ms to compute our SSC descriptor for one contour fragment, and $1.1$ s to encode the BSCP feature vector for one shape. The whole training process takes about $8$ hours (including feature computation and codebook learning), the testing process for one shape takes $17.5$ ms (excluding feature computation).

We evaluate our method on several shape classification benchmark datasets, including the MPEG-7 dataset~\cite{Ref:Latecki00}, the Animal dataset~\cite{Ref:Bai09}, and the ETH-80 dataset~\cite{Ref:Leibe03}. To avoid the biases caused by randomness, such a procedure is repeated 10 times. Average classification accuracy and standard derivation are reported to evaluate the performance of different shape classification methods. In each round, we randomly select half of shapes in each class to train and use the rest shapes to evaluate for every dataset except the ETH-80 dataset. On the ETH-80 dataset, following the previous methods~\cite{Ref:Leibe03,Ref:Lin07,Ref:Daliri08,Ref:Daliri10,Ref:Wang14}, we use all shapes except the current one for training and use the current one for testing (Leave-one-out setting~\cite{Ref:Devijver82}). Experimental results and analysis are given in the rest of this section.
\subsection{Animal Dataset}
We firstly test our method on the Animal dataset which is introduced in~\cite{Ref:Bai09}. This dataset contains 2000 shapes divided into 20 kinds of animals, including cat, spider, leopard, etc. It is the most challenging shape dataset due to the large intra-class variations caused by view point change and various gestures of animals (as shown in Fig.~\ref{fig:animal}). We randomly choose 50 shapes per class for training and leave the rest 50 shapes for testing. The comparison between BSCP and other shape classification methods is demonstrated in Table.~\ref{tbl:animal}.

As shown in Table.~\ref{tbl:animal}, the proposed method achieves a classification accuracy at $89.04\%$ which significantly outperforms the previous state-of-the-art method, Contextual BOW~\cite{Ref:Bharath15}, by over $3.0\%$. This result proves that the introduction of the object thickness information extracted from skeletons indeed help shape recognition. Our method also performs much better than BCF+BSP~\cite{Ref:Shen14}, evidencing that our method which associates a shape contour with skeletal information in such a principal way is more effective than the previous method, which combines contour and skeleton implicitly according to the weights learned by SVM. The comparison between our method and BCF~\cite{Ref:Wang14}, directly shows that SSC descriptor can capture not only the geometric information of the object contour but also the object thickness information for a shape. The combination of such two kinds of complementary information leads to an improvement on resisting interference caused by intra-class variations.

\begin{table}[!htbp]
\centering
\caption{Classification accuracy comparison on Animal dataset~\cite{Ref:Bai09}}\label{tbl:animal}
\begin{tabular}{cc}
\toprule
Algorithm&Classification accuracy\\
\midrule
Skeleton Paths~\cite{Ref:Bai09}&67.90\%\\
Contour Segments~\cite{Ref:Bai09}&71.70\%\\
IDSC~\cite{Ref:Lin07}&73.60\%\\
ICS~\cite{Ref:Bai09}&78.40\%\\
BCF~\cite{Ref:Wang14}&83.40 $\pm$ 1.30\%\\
Bioinformatic~\cite{Ref:Bicego15}&83.70\%\\
ShapeVocabulary~\cite{Ref:Bai14}&84.30 $\pm$ 1.01\%\\ 
BCF+BSP~\cite{Ref:Shen14}&85.50 $\pm$ 0.88\%\\
Contextual BOW~\cite{Ref:Bharath15}&86.00\%\\
\textbf{BSCP}&\textbf{89.04 $\pm$ 0.95\%}\\
\bottomrule
\end{tabular}
\end{table}

\begin{figure}[!h]
\centering
\includegraphics[width=1.0\linewidth]{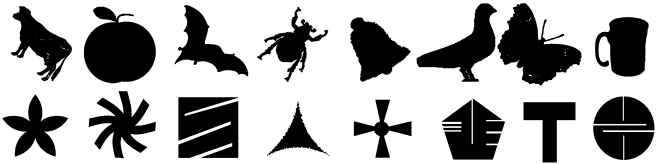}
\caption{Typical shapes of some classes from MPEG-7 dataset~\cite{Ref:Latecki00}.} \label{fig:mpeg7}
\end{figure}

\subsection{MPEG-7 Dataset}
Then we evaluate our method on the MPEG-7 dataset~\cite{Ref:Latecki00}, which is the most well-known dataset for shape analysis in the field of computer vision (see Fig.~\ref{fig:mpeg7}). 1400 images of the dataset are divided into 70 classes with high shape variability, in each of which there are 20 different shapes. Average classification accuracy and standard derivation of classification accuracies are reported in Table.~\ref{tbl:mpeg7}.

As shown in Table.~\ref{tbl:mpeg7}, our method achieves the best performance on the MPEG-7 dataset. BCF~\cite{Ref:Wang14} has already obtained good result, since it applies the Bag of Features framework to obtain the mid-level model of shape representation, which is more robust and accurate. BCF+BSP~\cite{Ref:Shen14} combines skeleton and contour information in a simple but effective way, and performs better than BCF, which proves that both skeleton and contour features are important in shape classification. However, with adopting SSC descriptor to combine contour and skeleton information, our method achieves better result than BCF+BSP on this dataset. The improvement on this dataset is not so significant as the one on the Animal dataset, the reason is the accuracies of the state-of-the-arts on this dataset have already approached to $100\%$.

\begin{table}[!htbp]
\centering
\caption{Classification accuracy comparison on MPEG-7 dataset~\cite{Ref:Latecki00}}\label{tbl:mpeg7}
\begin{tabular}{cc}
\toprule
Algorithm&Classification accuracy\\
\midrule
Skeleton Paths~\cite{Ref:Bai09}&86.70\%\\
Contour Segments~\cite{Ref:Bai09}&90.90\%\\
Bioinformatic~\cite{Ref:Bicego15}&96.10\%\\
ICS~\cite{Ref:Bai09}&96.60\%\\
BCF~\cite{Ref:Wang14}&97.16 $\pm$ 0.79\%\\
BCF+BSP~\cite{Ref:Shen14}&98.35 $\pm$ 0.63\%\\
\textbf{BSCP}&\textbf{98.41 $\pm$ 0.44\%}\\
\bottomrule
\end{tabular}
\end{table}

\subsection{ETH-80 Dataset}
The ETH-80 dataset~\cite{Ref:Leibe03} contains 80 objects, which are divided into 8 categories. There are 41 3-D color photographs token from different viewpoints for each object. We use the segmentation masks provided by the dataset to evaluate our method. The result is shown in Table.~\ref{tbl:eth80}.

Compared with other methods, ours achieves the classification accuracy of 93.05\%, outperforming the previous state-of-the-art approach in~\cite{Ref:Wang14} by over 1.5\%.

\begin{table}[!htbp]
\centering
\caption{Classification accuracy comparison on ETH-80 dataset~\cite{Ref:Leibe03}}\label{tbl:eth80}
\begin{tabular}{cc}
\toprule
Algorithm&Classification accuracy\\
\midrule
Color histogram~\cite{Ref:Leibe03}&64.86\%\\
PCA gray~\cite{Ref:Leibe03}&82.99\%\\
PCA masks~\cite{Ref:Leibe03}&83.41\%\\
SC+DP~\cite{Ref:Leibe03}&86.40\%\\
IDSC+DP~\cite{Ref:Lin07}&88.11\%\\
Robust symbolic~\cite{Ref:Daliri08}&90.28\%\\
Kernel-edit~\cite{Ref:Daliri10}&91.33\%\\
BCF~\cite{Ref:Wang14}&91.49\%\\
Bioinformatic~\cite{Ref:Bicego15}&91.50\%\\
\textbf{BSCP}&\textbf{93.05\%}\\
\bottomrule
\end{tabular}
\end{table}
\subsection{Parameter Discussion}
In this section, we investigate the effects of
three important parameters on shape classification accuracy.\\
\textbf{The number of object thickness difference bins for computing SSC.}
Since the proposal of the shape descriptor SSC is an important contribution, it is necessary to study how different settings of the descriptor effect the performance on shape classification.

\begin{figure}[!h]
\centering
\includegraphics[width=1.0\linewidth]{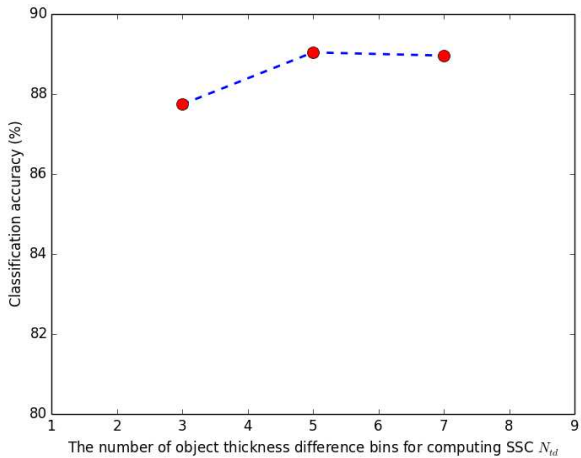}
\caption{Classification accuracies on Animal dataset~\cite{Ref:Bai09} by varying the number of object thickness difference bins for computing SSC $N_{td}$.} \label{fig:bins}
\end{figure}

As an extension of the Shape Context, SSC has one more dimension to describe the thickness differences, the number of object thickness difference bins $N_{td}$. To investigate the influence of this parameter, we set $N_{td}$ to different values to observe the performance change on the Animal dataset, while other parameters are set to the default values. The result is reported in Fig.~\ref{fig:bins}.

\begin{figure}[!h]
\centering
\includegraphics[width=1.0\linewidth]{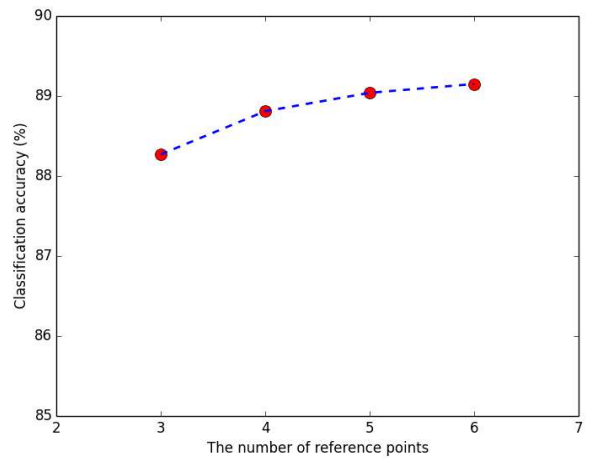}
\caption{Classification accuracies on Animal dataset~\cite{Ref:Bai09} by varying the number of reference points for computing SSC.} \label{fig:ref}
\end{figure}

\begin{figure}[!h]
\centering
\includegraphics[width=1.0\linewidth]{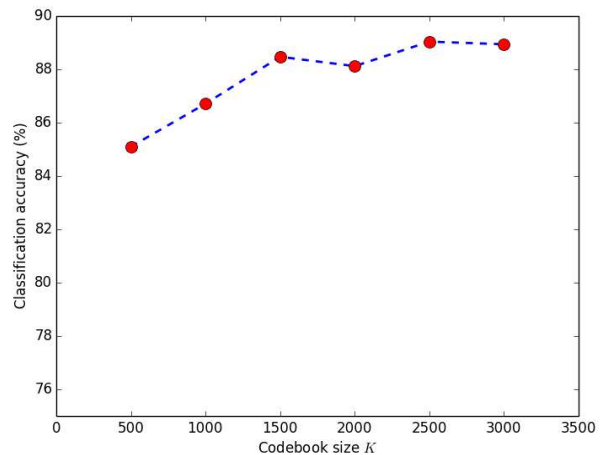}
\caption{Classification accuracies on Animal dataset~\cite{Ref:Bai09} by varying codebook size $K$.} \label{fig:codebook}
\end{figure}

Observed that our method achieves the best performance when $N_{td}$ is set to 5. $N_{td}=3$ (or $N_{td}=1$) leads to performance decrease. The reason may be that SSC with small $N_{td}$ can only give a coarse representation of the thickness information, while losing most of the information a skeleton provides. Although $N_{td}=7$ leads to a result close to the best one, it will result in significant increase in SSC descriptor computation, codebook learning and feature encoding. $N_{td}=5$, which is selected by us, is thought to be the best trade-off between accuracy and efficiency. We use it as the default value in our experiments, and gain the state-of-the-art performances on several datasets (see Table.~\ref{tbl:animal}, Table.~\ref{tbl:mpeg7} and Table.~\ref{tbl:eth80}).
\\
\textbf{The number of reference points for computing SSC.}
We also show how performance changes by varying the number of reference points when computing our SSC descriptor in Fig.~\ref{fig:ref}. Unsurprisingly, with the increase of the number of reference points, the classification accuracy is improved, as more shape details are considered. However, using more reference points leads to a significantly time consuming shape feature computation process. To balance the performance and computational cost, we choose $5$ reference points.
\\
\textbf{Codebook size.}
In this experiment, we adopt codebooks with different sizes, including 500, 1000, 1500, 2000, 2500 and 3000, to classify shapes on the Animal dataset. Other parameters are fixed to their default values. The classification accuracies of BSCP by using different codebook sizes are shown in the Fig.~\ref{fig:codebook}. As the codebook size increases, shape classification accuracy improves generally, which was also reported in~\cite{Ref:Wang14}.
\subsection{Limitation}
\begin{figure}[!h]
\centering
\includegraphics[width=1.0\linewidth]{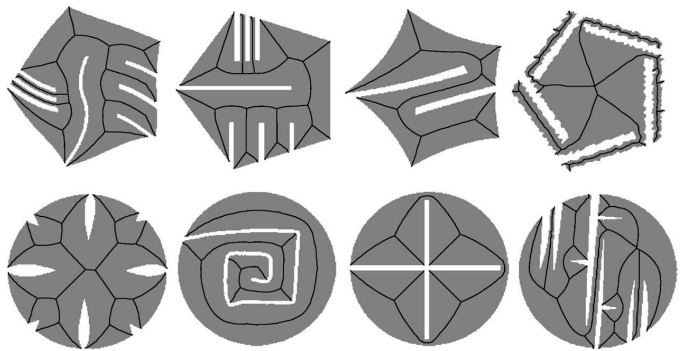}
\caption{Each row represents four shape examples from one kind of ``device'' class in the MPEG-7 dataset. The skeleton of each shape is visualized by black curves. The envelope contour of the shapes in each row are similar, while their skeletons are totally different.} \label{fig:limitation}
\end{figure}

Our SSC descriptor relies on the quality of the extracted skeleton. It also requires that the object can be well represented by its skeleton. Some objects in the MPEG-7 dataset, such as the ``device'' classes shown in Fig.~\ref{fig:limitation}, are not suitable to be represented by skeletons. In this case, our SSC descriptor does not perform well. We have applied our SSC descriptor to the shape retrieval framework of ``Shape Vocabulary''~\cite{Ref:Bai14} and test it on the MPEG-7 dataset. Unfortunately, we do not see the performance increase. This may be another reason why our method does not achieve an obvious classification improvement on the MPEG-7 dataset, as shown in Table~\ref{tbl:mpeg7}.

\section{Conclusion} \label{sec:con}
In this paper, we present a novel shape representation called BSCP, which combines contour and skeleton in a principal way.  This is achieved through the adoption of a novel low-level shape descriptor, the SSC, which is able to make full use of the natural correspondence between a contour and its skeleton. Both the normalization step and SPM are adopted to ensure that our method is effective and accurate, without losing the invariance to rotation. We have tested BSCP in many benchmarks, and the results lead to a conclusion that our method has achieved the state-of-the-art performance. Parameter discussion is also done as a reference for other researchers. In the future, we will further study how to apply BSCP to recognize objects in natural images, which requires reliable object contour detection~\cite{Ref:Shen15} and symmetry detection~\cite{Ref:Shen16}.

\noindent {\bf Acknowledgement}. This work was supported in part by the National Natural Science Foundation of China under Grant 61303095, in part by Research Fund for the Doctoral Program of Higher Education of China under Grant 20133108120017, in part by Innovation Program of Shanghai Municipal Education Commission under Grant 14YZ018, in part by Innovation Program of Shanghai University under Grant SDCX2013012 and in part by Cultivation Fund for the Young Faculty of Higher Education of Shanghai under Grant ZZSD13005.

\bibliographystyle{model2-names}
\bibliography{refs}

\begin{thebibliography}{46}
\expandafter\ifx\csname natexlab\endcsname\relax\def\natexlab#1{#1}\fi
\providecommand{\url}[1]{\texttt{#1}}
\providecommand{\href}[2]{#2}
\providecommand{\path}[1]{#1}
\providecommand{\DOIprefix}{doi:}
\providecommand{\ArXivprefix}{arXiv:}
\providecommand{\URLprefix}{URL: }
\providecommand{\Pubmedprefix}{pmid:}
\providecommand{\doi}[1]{\href{http://dx.doi.org/#1}{\path{#1}}}
\providecommand{\Pubmed}[1]{\href{pmid:#1}{\path{#1}}}
\providecommand{\bibinfo}[2]{#2}
\ifx\xfnm\relax \def\xfnm[#1]{\unskip,\space#1}\fi
\bibitem[{Aslan et~al.(2008)Aslan, Erdem, Erdem and Tari}]{Ref:Aslan08}
\bibinfo{author}{Aslan, C.}, \bibinfo{author}{Erdem, A.},
  \bibinfo{author}{Erdem, E.}, \bibinfo{author}{Tari, S.},
  \bibinfo{year}{2008}.
\newblock \bibinfo{title}{Disconnected skeleton: shape at its absolute scale}.
\newblock \bibinfo{journal}{IEEE Trans. Pattern Analysis and Machine
  Intelligence} \bibinfo{volume}{30}, \bibinfo{pages}{2188--2203}.
\bibitem[{Bai et~al.(2015)Bai, Bai, Zhu and Latecki}]{Ref:Bai15}
\bibinfo{author}{Bai, X.}, \bibinfo{author}{Bai, S.}, \bibinfo{author}{Zhu,
  Z.}, \bibinfo{author}{Latecki, L.J.}, \bibinfo{year}{2015}.
\newblock \bibinfo{title}{3d shape matching via two layer coding}.
\newblock \bibinfo{journal}{{IEEE} Trans. Pattern Anal. Mach. Intell.}
  \bibinfo{volume}{37}, \bibinfo{pages}{2361--2373}.
\bibitem[{Bai and Latecki(2008)}]{Ref:Bai08}
\bibinfo{author}{Bai, X.}, \bibinfo{author}{Latecki, L.}, \bibinfo{year}{2008}.
\newblock \bibinfo{title}{Path similarity skeleton graph matching}.
\newblock \bibinfo{journal}{IEEE Trans. Pattern Analysis and Machine
  Intelligence} \bibinfo{volume}{30}, \bibinfo{pages}{1282--1292}.
\bibitem[{Bai et~al.(2007)Bai, Latecki and Liu}]{Ref:Bai07}
\bibinfo{author}{Bai, X.}, \bibinfo{author}{Latecki, L.J.},
  \bibinfo{author}{Liu, W.}, \bibinfo{year}{2007}.
\newblock \bibinfo{title}{Skeleton pruning by contour partitioning with
  discrete curve evolution}.
\newblock \bibinfo{journal}{{IEEE} Trans. Pattern Anal. Mach. Intell.}
  \bibinfo{volume}{29}, \bibinfo{pages}{449--462}.
\bibitem[{Bai et~al.(2009)Bai, Liu and Tu}]{Ref:Bai09}
\bibinfo{author}{Bai, X.}, \bibinfo{author}{Liu, W.}, \bibinfo{author}{Tu, Z.},
  \bibinfo{year}{2009}.
\newblock \bibinfo{title}{Integrating contour and skeleton for shape
  classification}, in: \bibinfo{booktitle}{ICCV Workshops}, pp.
  \bibinfo{pages}{360--367}.
\bibitem[{Bai et~al.(2014)Bai, Rao and Wang}]{Ref:Bai14}
\bibinfo{author}{Bai, X.}, \bibinfo{author}{Rao, C.}, \bibinfo{author}{Wang,
  X.}, \bibinfo{year}{2014}.
\newblock \bibinfo{title}{Shape vocabulary: {A} robust and efficient shape
  representation for shape matching}.
\newblock \bibinfo{journal}{{IEEE} Transactions on Image Processing}
  \bibinfo{volume}{23}, \bibinfo{pages}{3935--3949}.
\bibitem[{Baseski et~al.(2009)Baseski, Erdem and Tari}]{Ref:Baseski09}
\bibinfo{author}{Baseski, E.}, \bibinfo{author}{Erdem, A.},
  \bibinfo{author}{Tari, S.}, \bibinfo{year}{2009}.
\newblock \bibinfo{title}{Dissimilarity between two skeletal trees in a
  context}.
\newblock \bibinfo{journal}{Pattern Recognition} \bibinfo{volume}{42},
  \bibinfo{pages}{370--385}.
\bibitem[{Belongie et~al.(2002)Belongie, Malik and Puzicha}]{Ref:Belongie02}
\bibinfo{author}{Belongie, S.}, \bibinfo{author}{Malik, J.},
  \bibinfo{author}{Puzicha, J.}, \bibinfo{year}{2002}.
\newblock \bibinfo{title}{Shape matching and object recognition using shape
  contexts}.
\newblock \bibinfo{journal}{IEEE Trans. Pattern Analysis and Machine
  Intelligence} \bibinfo{volume}{24}, \bibinfo{pages}{509--522}.
\bibitem[{Bharath et~al.(2015)Bharath, Xiang and Lee}]{Ref:Bharath15}
\bibinfo{author}{Bharath, R.}, \bibinfo{author}{Xiang, C.},
  \bibinfo{author}{Lee, T.H.}, \bibinfo{year}{2015}.
\newblock \bibinfo{title}{Shape classification using invariant features and
  contextual information in the bag-of-words model}.
\newblock \bibinfo{journal}{Pattern Recognition} \bibinfo{volume}{48},
  \bibinfo{pages}{894--906}.
\bibitem[{Bicego and Lovato(2015)}]{Ref:Bicego15}
\bibinfo{author}{Bicego, M.}, \bibinfo{author}{Lovato, P.},
  \bibinfo{year}{2015}.
\newblock \bibinfo{title}{A bioinformatics approach to 2d shape
  classification}.
\newblock \bibinfo{journal}{Computer Vision and Image Understanding} .
\bibitem[{Blum(1973)}]{Ref:Blum73}
\bibinfo{author}{Blum, H.}, \bibinfo{year}{1973}.
\newblock \bibinfo{title}{Biological shape and visual science}.
\newblock \bibinfo{journal}{J. Theor. Biol.} \bibinfo{volume}{38},
  \bibinfo{pages}{205--287}.
\bibitem[{Borgefors et~al.(1999)Borgefors, Nystr{\"{o}}m and
  di~Baja}]{Ref:Borgefors99}
\bibinfo{author}{Borgefors, G.}, \bibinfo{author}{Nystr{\"{o}}m, I.},
  \bibinfo{author}{di~Baja, G.S.}, \bibinfo{year}{1999}.
\newblock \bibinfo{title}{Computing skeletons in three dimensions}.
\newblock \bibinfo{journal}{Pattern Recognition} \bibinfo{volume}{32},
  \bibinfo{pages}{1225--1236}.
\bibitem[{Cormen et~al.(2001)Cormen, Leiserson, Rivest and
  Stein}]{Ref:Cormen01}
\bibinfo{author}{Cormen, T.}, \bibinfo{author}{Leiserson, C.},
  \bibinfo{author}{Rivest, R.}, \bibinfo{author}{Stein, C.},
  \bibinfo{year}{2001}.
\newblock \bibinfo{title}{Introduction to Algorithms, second ed.}
\newblock \bibinfo{publisher}{MIT Press}.
\bibitem[{Crammer and Singer(2001)}]{Ref:Crammer01}
\bibinfo{author}{Crammer, K.}, \bibinfo{author}{Singer, Y.},
  \bibinfo{year}{2001}.
\newblock \bibinfo{title}{On the algorithmic implementation of multiclass
  kernel-based vector machines}.
\newblock \bibinfo{journal}{Journal of Machine Learning Research}
  \bibinfo{volume}{2}, \bibinfo{pages}{265--292}.
\bibitem[{Daliri and Torre(2008)}]{Ref:Daliri08}
\bibinfo{author}{Daliri, M.R.}, \bibinfo{author}{Torre, V.},
  \bibinfo{year}{2008}.
\newblock \bibinfo{title}{Robust symbolic representation for shape recognition
  and retrieval}.
\newblock \bibinfo{journal}{Pattern Recognition} \bibinfo{volume}{41},
  \bibinfo{pages}{1782--1798}.
\bibitem[{Daliri and Torre(2010)}]{Ref:Daliri10}
\bibinfo{author}{Daliri, M.R.}, \bibinfo{author}{Torre, V.},
  \bibinfo{year}{2010}.
\newblock \bibinfo{title}{Shape recognition based on kernel-edit distance}.
\newblock \bibinfo{journal}{Computer Vision and Image Understanding}
  \bibinfo{volume}{114}, \bibinfo{pages}{1097--1103}.
\bibitem[{Demirci et~al.(2006)Demirci, Shokoufandeh, Keselman, Bretzner and
  Dickinson}]{Ref:Demirci06}
\bibinfo{author}{Demirci, M.}, \bibinfo{author}{Shokoufandeh, A.},
  \bibinfo{author}{Keselman, Y.}, \bibinfo{author}{Bretzner, L.},
  \bibinfo{author}{Dickinson, S.}, \bibinfo{year}{2006}.
\newblock \bibinfo{title}{Object recognition as many-to-many feature matching}.
\newblock \bibinfo{journal}{Int'l J. Computer Vision} \bibinfo{volume}{69},
  \bibinfo{pages}{203--222}.
\bibitem[{Devijver and Kittler(1982)}]{Ref:Devijver82}
\bibinfo{author}{Devijver, P.A.}, \bibinfo{author}{Kittler, J.},
  \bibinfo{year}{1982}.
\newblock \bibinfo{title}{Pattern Recognition: A Statistical Approach}.
\newblock \bibinfo{publisher}{London, GB: Prentice-Hall}.
\bibitem[{Duchon(1977)}]{Ref:Duchon77}
\bibinfo{author}{Duchon, J.}, \bibinfo{year}{1977}.
\newblock \bibinfo{title}{Splines Minimizing Rotation-Invariant Semi-Norms in
  Sobolev Spaces}.
\newblock \bibinfo{publisher}{Berlin: Springer-Verlag}.
\bibitem[{Erdem and Tari(2010)}]{Ref:Erdem10}
\bibinfo{author}{Erdem, A.}, \bibinfo{author}{Tari, S.}, \bibinfo{year}{2010}.
\newblock \bibinfo{title}{A similarity-based approach for shape classification
  using aslan skeletons}.
\newblock \bibinfo{journal}{Pattern Recognition Letters} \bibinfo{volume}{31},
  \bibinfo{pages}{2024--2032}.
\bibitem[{Felzenszwalb and Schwartz(2007)}]{Ref:Felzenszwalb07}
\bibinfo{author}{Felzenszwalb, P.F.}, \bibinfo{author}{Schwartz, J.},
  \bibinfo{year}{2007}.
\newblock \bibinfo{title}{Hierarchical matching of deformable shapes}, in:
  \bibinfo{booktitle}{CVPR}.
\bibitem[{Grigorescu and Petkov(2003)}]{Ref:Grigorescu03}
\bibinfo{author}{Grigorescu, C.}, \bibinfo{author}{Petkov, N.},
  \bibinfo{year}{2003}.
\newblock \bibinfo{title}{Distance sets for shape filters and shape
  recognition}.
\newblock \bibinfo{journal}{IEEE Transactions on Image Processing}
  \bibinfo{volume}{12}, \bibinfo{pages}{1274--1286}.
\bibitem[{Latecki and Lak{\"{a}}mper(1999)}]{Ref:LateckiL99}
\bibinfo{author}{Latecki, L.J.}, \bibinfo{author}{Lak{\"{a}}mper, R.},
  \bibinfo{year}{1999}.
\newblock \bibinfo{title}{Convexity rule for shape decomposition based on
  discrete contour evolution}.
\newblock \bibinfo{journal}{Computer Vision and Image Understanding}
  \bibinfo{volume}{73}, \bibinfo{pages}{441--454}.
\bibitem[{Latecki et~al.(2000)Latecki, Lak{\"a}mper and
  Eckhardt}]{Ref:Latecki00}
\bibinfo{author}{Latecki, L.J.}, \bibinfo{author}{Lak{\"a}mper, R.},
  \bibinfo{author}{Eckhardt, U.}, \bibinfo{year}{2000}.
\newblock \bibinfo{title}{Shape descriptors for non-rigid shapes with a single
  closed contour}, in: \bibinfo{booktitle}{CVPR}, pp.
  \bibinfo{pages}{1424--1429}.
\bibitem[{Lazebnik et~al.(2006)Lazebnik, Schmid and Ponce}]{Ref:Lazebnik06}
\bibinfo{author}{Lazebnik, S.}, \bibinfo{author}{Schmid, C.},
  \bibinfo{author}{Ponce, J.}, \bibinfo{year}{2006}.
\newblock \bibinfo{title}{Beyond bags of features: Spatial pyramid matching for
  recognizing natural scene categories}, in: \bibinfo{booktitle}{CVPR}, pp.
  \bibinfo{pages}{2169--2178}.
\bibitem[{Leibe and Schiele(2003)}]{Ref:Leibe03}
\bibinfo{author}{Leibe, B.}, \bibinfo{author}{Schiele, B.},
  \bibinfo{year}{2003}.
\newblock \bibinfo{title}{Analyzing appearance and contour based methods for
  object categorization}, in: \bibinfo{booktitle}{2003 {IEEE} Computer Society
  Conference on Computer Vision and Pattern Recognition {(CVPR} 2003), 16-22
  June 2003, Madison, WI, {USA}}, pp. \bibinfo{pages}{409--415}.
\bibitem[{Ling and Jacobs(2007)}]{Ref:Lin07}
\bibinfo{author}{Ling, H.}, \bibinfo{author}{Jacobs, D.W.},
  \bibinfo{year}{2007}.
\newblock \bibinfo{title}{Shape classification using the inner-distance}.
\newblock \bibinfo{journal}{IEEE Trans. Pattern Analysis and Machine
  Intelligence} \bibinfo{volume}{29}, \bibinfo{pages}{286--299}.
\bibitem[{Ma et~al.(2015a)Ma, Qiu, Zhao, Ma, Yuille and Tu}]{Ref:Ma15}
\bibinfo{author}{Ma, J.}, \bibinfo{author}{Qiu, W.}, \bibinfo{author}{Zhao,
  J.}, \bibinfo{author}{Ma, Y.}, \bibinfo{author}{Yuille, A.L.},
  \bibinfo{author}{Tu, Z.}, \bibinfo{year}{2015}a.
\newblock \bibinfo{title}{Robust l\({}_{\mbox{2}}\)e estimation of
  transformation for non-rigid registration}.
\newblock \bibinfo{journal}{{IEEE} Transactions on Signal Processing}
  \bibinfo{volume}{63}, \bibinfo{pages}{1115--1129}.
\bibitem[{Ma et~al.(2014)Ma, Zhao, Tian, Yuille and Tu}]{Ref:Ma14}
\bibinfo{author}{Ma, J.}, \bibinfo{author}{Zhao, J.}, \bibinfo{author}{Tian,
  J.}, \bibinfo{author}{Yuille, A.L.}, \bibinfo{author}{Tu, Z.},
  \bibinfo{year}{2014}.
\newblock \bibinfo{title}{Robust point matching via vector field consensus}.
\newblock \bibinfo{journal}{IEEE Trans. Image Process.} \bibinfo{volume}{23},
  \bibinfo{pages}{1706--1721}.
\bibitem[{Ma et~al.(2016)Ma, Zhao and Yuille}]{Ref:Ma16}
\bibinfo{author}{Ma, J.}, \bibinfo{author}{Zhao, J.}, \bibinfo{author}{Yuille,
  A.L.}, \bibinfo{year}{2016}.
\newblock \bibinfo{title}{Non-rigid point set registration by preserving global
  and local structures}.
\newblock \bibinfo{journal}{IEEE Transactions on Image Processing}
  \bibinfo{volume}{25}, \bibinfo{pages}{53--64}.
\bibitem[{Ma et~al.(2015b)Ma, Zhou, Zhao and Tian}]{Ref:Ma15d}
\bibinfo{author}{Ma, J.}, \bibinfo{author}{Zhou, H.}, \bibinfo{author}{Zhao,
  J.}, \bibinfo{author}{Tian, J.}, \bibinfo{year}{2015}b.
\newblock \bibinfo{title}{Robust feature matching for remote sensing image
  registration via locally linear transforming}.
\newblock \bibinfo{journal}{IEEE Transactions on Geoscience and Remote Sensing}
  \bibinfo{volume}{53}, \bibinfo{pages}{6469--6481}.
\bibitem[{Macrini et~al.(2011)Macrini, Dickinson, Fleet and
  Siddiqi}]{Ref:Macrini11}
\bibinfo{author}{Macrini, D.}, \bibinfo{author}{Dickinson, S.J.},
  \bibinfo{author}{Fleet, D.J.}, \bibinfo{author}{Siddiqi, K.},
  \bibinfo{year}{2011}.
\newblock \bibinfo{title}{Object categorization using bone graphs}.
\newblock \bibinfo{journal}{Computer Vision and Image Understanding}
  \bibinfo{volume}{115}, \bibinfo{pages}{1187--1206}.
\bibitem[{Saha et~al.(2015)Saha, Borgefors and di~Baja}]{Ref:Saha15}
\bibinfo{author}{Saha, P.K.}, \bibinfo{author}{Borgefors, G.},
  \bibinfo{author}{di~Baja, G.S.}, \bibinfo{year}{2015}.
\newblock \bibinfo{title}{A survey on skeletonization algorithms and their
  applications}.
\newblock \bibinfo{journal}{Pattern Recognition Letters} .
\bibitem[{Sebastian et~al.(2004)Sebastian, Klein and Kimia}]{Ref:Sebastian04}
\bibinfo{author}{Sebastian, T.}, \bibinfo{author}{Klein, P.},
  \bibinfo{author}{Kimia, B.}, \bibinfo{year}{2004}.
\newblock \bibinfo{title}{Recognition of shapes by editing their shock graphs}.
\newblock \bibinfo{journal}{IEEE Trans. Pattern Analysis and Machine
  Intelligence} \bibinfo{volume}{26}, \bibinfo{pages}{550--571}.
\bibitem[{Shen et~al.(2011)Shen, Bai, Hu, Wang and Latecki}]{Ref:Shen11}
\bibinfo{author}{Shen, W.}, \bibinfo{author}{Bai, X.}, \bibinfo{author}{Hu,
  R.}, \bibinfo{author}{Wang, H.}, \bibinfo{author}{Latecki, L.J.},
  \bibinfo{year}{2011}.
\newblock \bibinfo{title}{Skeleton growing and pruning with bending potential
  ratio}.
\newblock \bibinfo{journal}{Pattern Recognition} \bibinfo{volume}{44},
  \bibinfo{pages}{196--209}.
\bibitem[{Shen et~al.(2016)Shen, Bai, Hu and Zhang}]{Ref:Shen16}
\bibinfo{author}{Shen, W.}, \bibinfo{author}{Bai, X.}, \bibinfo{author}{Hu,
  Z.}, \bibinfo{author}{Zhang, Z.}, \bibinfo{year}{2016}.
\newblock \bibinfo{title}{Multiple instance subspace learning via partial
  random projection tree for local reflection symmetry in natural images}.
\newblock \bibinfo{journal}{Pattern Recognition} .
\bibitem[{Shen et~al.(2013a)Shen, Bai, Yang and Latecki}]{Ref:Shen13}
\bibinfo{author}{Shen, W.}, \bibinfo{author}{Bai, X.}, \bibinfo{author}{Yang,
  X.}, \bibinfo{author}{Latecki, L.J.}, \bibinfo{year}{2013}a.
\newblock \bibinfo{title}{Skeleton pruning as trade-off between skeleton
  simplicity and reconstruction error}.
\newblock \bibinfo{journal}{SCIENCE CHINA Information Sciences}
  \bibinfo{volume}{56}, \bibinfo{pages}{1--14}.
\bibitem[{Shen et~al.(2015)Shen, Wang, Wang, Bai and Zhang}]{Ref:Shen15}
\bibinfo{author}{Shen, W.}, \bibinfo{author}{Wang, X.}, \bibinfo{author}{Wang,
  Y.}, \bibinfo{author}{Bai, X.}, \bibinfo{author}{Zhang, Z.},
  \bibinfo{year}{2015}.
\newblock \bibinfo{title}{Deepcontour: {A} deep convolutional feature learned
  by positive-sharing loss for contour detection}, in:
  \bibinfo{booktitle}{{IEEE} Conference on Computer Vision and Pattern
  Recognition, {CVPR} 2015, Boston, MA, USA, June 7-12, 2015}, pp.
  \bibinfo{pages}{3982--3991}.
\bibitem[{Shen et~al.(2014)Shen, Wang, Yao and Bai}]{Ref:Shen14}
\bibinfo{author}{Shen, W.}, \bibinfo{author}{Wang, X.}, \bibinfo{author}{Yao,
  C.}, \bibinfo{author}{Bai, X.}, \bibinfo{year}{2014}.
\newblock \bibinfo{title}{Shape recognition by combining contour and skeleton
  into a mid-level representation}, in: \bibinfo{booktitle}{Pattern Recognition
  - 6th Chinese Conference, {CCPR} 2014, Changsha, China, November 17-19, 2014.
  Proceedings, Part {I}}, pp. \bibinfo{pages}{391--400}.
\bibitem[{Shen et~al.(2013b)Shen, Wang, Bai, Wang and Latecki}]{Ref:ShenPR13}
\bibinfo{author}{Shen, W.}, \bibinfo{author}{Wang, Y.}, \bibinfo{author}{Bai,
  X.}, \bibinfo{author}{Wang, H.}, \bibinfo{author}{Latecki, L.J.},
  \bibinfo{year}{2013}b.
\newblock \bibinfo{title}{Shape clustering: Common structure discovery}.
\newblock \bibinfo{journal}{Pattern Recognition} \bibinfo{volume}{46},
  \bibinfo{pages}{539--550}.
\bibitem[{Siddiqi et~al.(1999)Siddiqi, Shokoufandeh, Dickinson and
  Zucker}]{Ref:Siddiqi99}
\bibinfo{author}{Siddiqi, K.}, \bibinfo{author}{Shokoufandeh, A.},
  \bibinfo{author}{Dickinson, S.}, \bibinfo{author}{Zucker, S.},
  \bibinfo{year}{1999}.
\newblock \bibinfo{title}{Shock graphs and shape matching}.
\newblock \bibinfo{journal}{Int'l J. Computer Vision} \bibinfo{volume}{35},
  \bibinfo{pages}{13--32}.
\bibitem[{Sun and Super(2005)}]{Ref:Sun05}
\bibinfo{author}{Sun, K.B.}, \bibinfo{author}{Super, B.J.},
  \bibinfo{year}{2005}.
\newblock \bibinfo{title}{Classification of contour shapes using class segment
  sets}, in: \bibinfo{booktitle}{CVPR}, pp. \bibinfo{pages}{727--733}.
\bibitem[{Wang et~al.(2010a)Wang, Shen, Liu, You and Bai}]{Ref:WangB10}
\bibinfo{author}{Wang, B.}, \bibinfo{author}{Shen, W.}, \bibinfo{author}{Liu,
  W.}, \bibinfo{author}{You, X.}, \bibinfo{author}{Bai, X.},
  \bibinfo{year}{2010}a.
\newblock \bibinfo{title}{Shape classification using tree -unions}, in:
  \bibinfo{booktitle}{ICPR}, pp. \bibinfo{pages}{983--986}.
\bibitem[{Wang et~al.(2010b)Wang, Yang, Yu, Lv, Huang and Gong}]{Ref:Wang10}
\bibinfo{author}{Wang, J.}, \bibinfo{author}{Yang, J.}, \bibinfo{author}{Yu,
  K.}, \bibinfo{author}{Lv, F.}, \bibinfo{author}{Huang, T.S.},
  \bibinfo{author}{Gong, Y.}, \bibinfo{year}{2010}b.
\newblock \bibinfo{title}{Locality-constrained linear coding for image
  classification}, in: \bibinfo{booktitle}{CVPR}, pp.
  \bibinfo{pages}{3360--3367}.
\bibitem[{Wang et~al.(2014)Wang, Feng, Bai, Liu and Latecki}]{Ref:Wang14}
\bibinfo{author}{Wang, X.}, \bibinfo{author}{Feng, B.}, \bibinfo{author}{Bai,
  X.}, \bibinfo{author}{Liu, W.}, \bibinfo{author}{Latecki, L.J.},
  \bibinfo{year}{2014}.
\newblock \bibinfo{title}{Bag of contour fragments for robust shape
  classification}.
\newblock \bibinfo{journal}{Pattern Recognition} \bibinfo{volume}{47},
  \bibinfo{pages}{2116--2125}.
\bibitem[{Xie et~al.(2008)Xie, Heng and Shah}]{Ref:Xie08}
\bibinfo{author}{Xie, J.}, \bibinfo{author}{Heng, P.}, \bibinfo{author}{Shah,
  M.}, \bibinfo{year}{2008}.
\newblock \bibinfo{title}{Shape matching and modeling using skeletal context}.
\newblock \bibinfo{journal}{Pattern Recognition} \bibinfo{volume}{41},
  \bibinfo{pages}{1756--1767}.

\end{thebibliography}

\end{document}